\documentclass[arxiv]{melba}

\usepackage{mwe} 

\usepackage{amsmath,amsfonts}
\usepackage{threeparttable}

\usepackage{booktabs}       
\usepackage{multirow}       
\usepackage{graphicx}       

\melbaid{YYYY:NNN}  
\doi{https://doi.org/10.59275/j.melba.2024-AAAA}
\melbaauthors{Yunzhi Xu}  
\volume{2}
\firstpageno{1337}  
\melbayear{YYYY}  
\datesubmitted{07/2025}  
\datepublished{m2/2025}  

\melbaspecialissue{Medical Imaging with Deep Learning (MIDL) 2025}
\melbaspecialissueeditors{Marleen de Bruijne, Tal Arbel, Ismail Ben Ayed, Hervé Lombaert}

\ShortHeadings{HyKid: ANNOTATED PEDIATRIC HYDROCEPHALUS}{Xu}

\title{HyKid: An Open MRI Dataset with Expert-Annotated Multi-Structure and Choroid Plexus in Pediatric Hydrocephalus}

\author{
	\firstname Yunzhi \surname Xu\aff{1},
	\name Yushuang Ding\aff{2}
	\name Hu Sun\aff{3}
	\name Hongxi Zhang \aff{2}
	\name Li Zhao\aff{1}
}
\affiliations{
	\num 1 \addr College of Biomedical Engineering and Instrument Science, Zhejiang University, Hangzhou, China. \\
	\num 2 \addr Department of Radiology, Children's Hospital, Zhejiang University School of Medicine, National Clinical Research Center for Child Health, Hangzhou, China \\
	\num 3 \addr Neurosurgery of Zhejiang Hospital, Hangzhou, China. \\  
}

\abstract{
Evaluation of hydrocephalus in children is challenging, and the related research is limited by a lack of publicly available, expert-annotated datasets, particularly those with segmentation of the choroid plexus. To address this, we present HyKid, an open-source dataset from 48 pediatric patients with hydrocephalus. 3D MRIs were provided with 1mm isotropic resolution, which was reconstructed from routine low-resolution images using a slice-to-volume algorithm. Manually corrected segmentations of brain tissues, including white matter, grey matter, lateral ventricle, external CSF, and the choroid plexus, were provided by an experienced neurologist. Additionally, structured data was extracted from clinical radiology reports using a Retrieval-Augmented Generation framework. The strong correlation between choroid plexus volume and total CSF volume provided a potential biomarker for hydrocephalus evaluation, achieving excellent performance in a predictive model (AUC \text{=} 0.87). The proposed HyKid dataset provided a high-quality benchmark for neuroimaging algorithms development, and it revealed the choroid plexus related features in hydrocephalus assessments. 
	Our datasets are publicly available at~\url{https://www.synapse.org/Synapse:syn68544889}.}


\begin{document}

\twocolumn[\maketitle]

\section{Introduction}
\subsection{Background}
    \begin{figure*}[h]
        \centering
        \includegraphics[width=0.7\linewidth, height=.3\textheight]{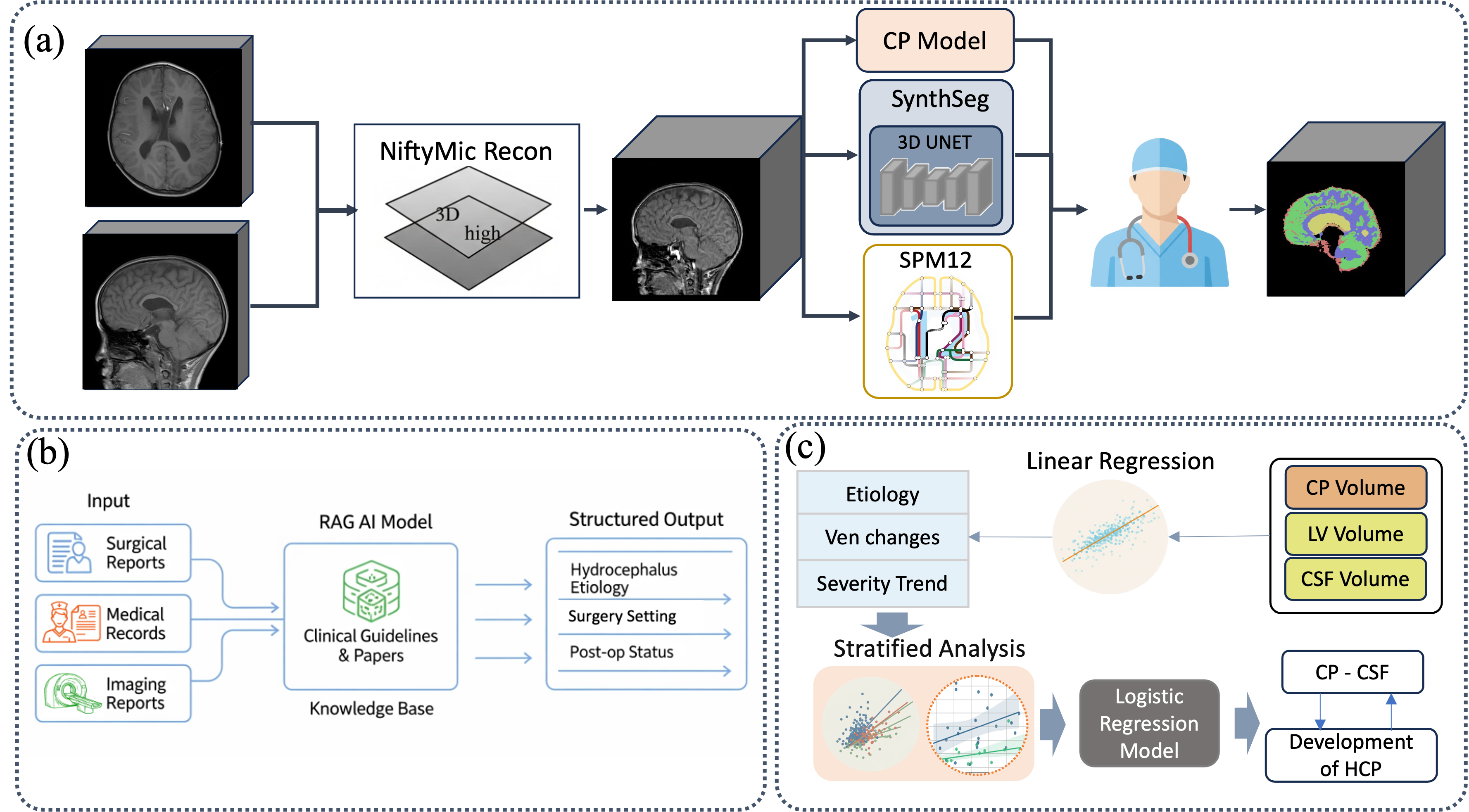}
        \caption{The methodological framework. (a) The pipeline for 3D high-resolution image reconstruction and semi-automated segmentation. (b) The Retrieval-Augmented Generation (RAG) framework for extracting structured clinical data from unstructured reports. (c) The statistical framework for analyzing the association between choroid plexus (CP) and lateral ventricles, and between CP and total CSF volume, incorporating stratified analysis, linear regression, ANCOVA, and logistic regression.}
		\label{fig:framework}
\end{figure*}

\enluminure{H}ydrocephalus is a serious condition characterized by the abnormal accumulation of cerebrospinal fluid (CSF), which can be caused by congenital malformations, trauma, or tumors \citep{laurenceNaturalHistoryHydrocephalus1962}. The CSF, mainly generated by the choroid plexus, flows through the interventricular foramen into the third ventricle, and then travels through the cerebral aqueduct to the fourth ventricle. From there, it passes through the median and lateral apertures into the subarachnoid space \citep{damkierCerebrospinalFluidSecretion2013}. Besides congenital hydrocephalus, acquired hydrocephalus can be caused by tumors, infections, hemorrhage, post-traumatic events, and idiopathic factors. Disruption of the CSF circulation equilibrium causes excessive CSF accumulation within the cranium, leading to hydrocephalus \citep{kahleHydrocephalusChildren2016a}. 

One common cause of hydrocephalus is the obstruction in the CSF pathway, which leads to CSF accumulation and elevated intracranial pressure. The incidence of pediatric hydrocephalus is approximately 1 in 500 births \citep{flanneryPediatricHydrocephalusSystematic2014},  which is the leading cause of brain surgery in children and which is often associated with severe postoperative sequelae \citep{kulkarniOutcomesCSFShunting2013}. Delayed or inadequate treatment of hydrocephalus can lead to brain damage, such as psychiatric and physical disabilities. For instance, Melot et al. found that 60.4\% of hydrocephalus children presented with fine motor dysfunction, impacting accuracy and dexterity, while 41.9\% had visual impairment \citep{melotNeurodevelopmentalLongtermOutcome2016}. Furthermore, psychiatric and behavioral follow-ups have documented learning disabilities and autism \citep{lindquist2006behavioural}.

Ventriculoperitoneal (VP) shunting is a standard treatment for pediatric hydrocephalus \citep{patwardhanImplantedVentricularShunts2005}, which creates a new flow-out path for CSF to relieve its accumulation. The procedure implants a shunt system with a one-way valve to drain excess CSF from the ventricles to the peritoneal cavity \citep{garegnaniVentriculoperitonealShuntingDevices2020}. However, this procedure is associated with well-known complications, including high rates of shunt malfunction and brain infection \citep{kestleStandardizedProtocolReduce2011,sedanoAssociationsStandardCare2023a}. Under- or over drainage of CSF can lead to a range of symptoms including headaches, nausea, vomiting, blurred vision, and dizziness \citep{rosShuntOverdrainageReappraisal2021}. However, the proper amount of CSF drainage is determined in a TAP exam typically, which requires lumbar puncture and is evaluated in a few days. Therefore, it is highly demanded to evaluate the hydrocephalus based on  image features to simplify the procedure and improve patient outcomes.

 The ventricular volume is a conventional image bioma-\\rker used in hydrocephalus evaluation \citep{rekateContemporaryDefinitionClassification2009}, which can be quantified in the Evans index \citep{evans1942encephalographic}, temporal horns angle, and subarachnoid value. Meanwhile, other image-based features could be used to evaluate hydrocephalus, but have not been explored thoroughly. For example, the choroid plexus (CP), the primary site of CSF production within the ventricles, may have morphological and functional abnormalities that are closely linked to the pathophysiology of CSF circulatory disorders and hydrocephalus. Therefore, quantitative analysis of the choroid plexus can provide essential indicators for understanding hydrocephalus pathophysiology. In addition, hydrocephalus could damages adjacent brain tissue, leading to periventricular white matter atrophy \citep{delbigioNeuropathologyStructuralChanges2010}\citep{mcallisterPathophysiologyCongenitalNeonatal2012} and the abnormality of the gray matter which is correlated with the cognitive function \citep{fletcher1992cerebral}. Therefore, its medical images may provide potential new features to evaluate the functional changes of hydrocephalus besides the conventional structural alterations.

\subsection{Existing Datasets and Methodologies:}
Although large-scale clinical data platforms like the UK Shunt Registry \citep{richards2009efficacy} and the Hydrocephalus Clinical Research Network (HCRN) \citep{kulkarniOutcomesCSFShunting2013}  exist, pediatric hydrocephalus datasets are seldom available to the public, especially well-annotated MRI \\datasets. 
On the other hand, pediatric brain MRIs were provided in open datasets \citep{akinci2023development}, but hydrocephalus cases were very rare. The primary reasons for the shortage of pediatric hydrocephalus data are motion artifacts in pediatric MRI and low image resolution resulting from fast acquisition sequences used to reduce the motion \citep{haOneMinuteUltrafastBrain2020}. Consequently, datasets featuring expert-derived, voxel-level ground-truth segmentations were infrequently available \citep{tahaAutomatedVentricularSegmentation2025}. In addition, the structured clinical information, including the changing trend of ventricles and clinical conditions, is seldom provided to validate the performance of image biomarkers, which hinders the developments of novel models of clinical evaluations and outcomes.
\begin{figure*}[h]
    \centering
    \includegraphics[width=1\linewidth]{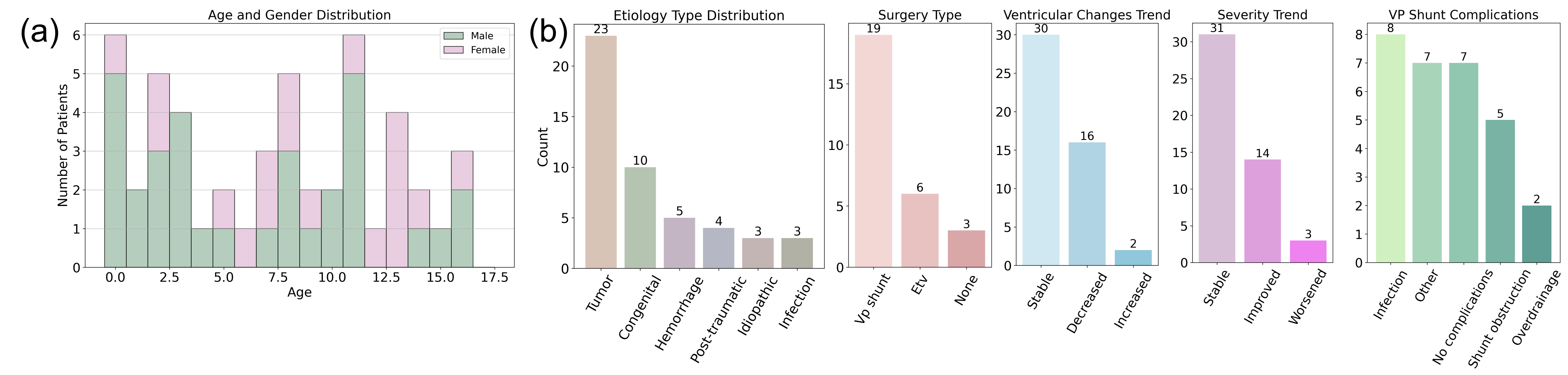}
    \caption{Demographic and clinical characteristics of the pediatric hydrocephalus dataset. (a) Patient distribution by age and gender. (b) Distribution of key clinical features.}
	\label{fig:demographics}
\end{figure*}

\begin{table}[htbp]
	\centering
	\caption{Demographic characteristics and MRI acquisition.} 
	\resizebox{0.48\textwidth}{!}{%
	\begin{tabular}{ll}
	  \toprule
	  \multicolumn{2}{c}{Statistics} \\ 
	\midrule
		  Cases & $48$ \\
		  Male/Female, n [\%] & $31 (64.6\%) / 17 (35.4\%)$ \\
		  Age, years, mean, std & $7.7364 \pm 5.0681$ \\
		  median [IQR] & $8.3435 [2.8388 - 11.4938]$ \\
		  Number of axial slices, n & $2054$ \\
		  Slice thickness, mm, mean [SD] & $7.1163 \pm 0.65615$ \\
		  Slices per case, mean [SD] & $19.0$ \\
            In-plane resolution, mm, mean [SD] & $0.4202 \pm 0.0384$\\
            In-plane matrix size & $512 \times 512$\\
	  \bottomrule
	\end{tabular}
	}
	\label{tab:Demographic} 
\end{table}

\subsection{Contributions}

MRI can provide great soft tissue contrast and reveal potential changes caused by hydrocephalus. High-resolution MRI has been employed in research, enabling more precise anatomical diagnoses of hydrocephalus through morphometric analysis and volumetric measurements \citep{hunt2003assessment}\citep{engelhardt2015regional}. The present work attempted to establish the dataset and features from high-resolution images and precise structural segmentations to enable the quantification of more subtle brain structural characteristics in pediatric hydrocephalus.

The contributions of this work are three-fold: First, this study provided HyKid dataset from 48 pediatric hydrocephalus patients with 2,054 slices, which includes: 1) 3D high-resolution images reconstructed using a slice-to-volume technique; 2) manual segmentations of gray matter, white matter, external CSF, lateral ventricles, and the choroid plexus, performed by an experienced neurologist; and 3) structured clinical records parsed using a knowledge retrieval-augmented technique. The pipeline of HyKid dataset is illustrated in Figure \ref{fig:framework}. Second, through a stratified clinical analysis, significant correlations between CP volume and the volumes of the ventricles and total CSF were found, which could provide an indicator of clinical improvement. Third, the total global CSF volume, rather than ventricular volume, is a superior metric for predicting clinical symptoms, which may reveal the link between the choroid plexus and CSF and the patient's clinical trajectory. This is evidenced by a 7\% increase in explained variance ($R^2$, ANCOVA) and a 0.07 improvement in predictive accuracy (AUC, logistic regression).

\section{Dataset Description}
This dataset was collected retrospectively under the Institutional Review Board of Approval  (No.20240520090500000\\6293). Before data sharing, both imaging and clinical text data were anonymized to protect patient privacy. Specifically, imaging data were de-identified by removing personally identifiable information tags from the DICOM headers and converting all files to the NifTI format. For clinical reports, patient names were removed before processing with the Retrieval-Augmented Generation (RAG) technique \citep{lewis2020retrieval}.

\begin{figure*}[h]
    \centering
    \includegraphics[width=1\linewidth]{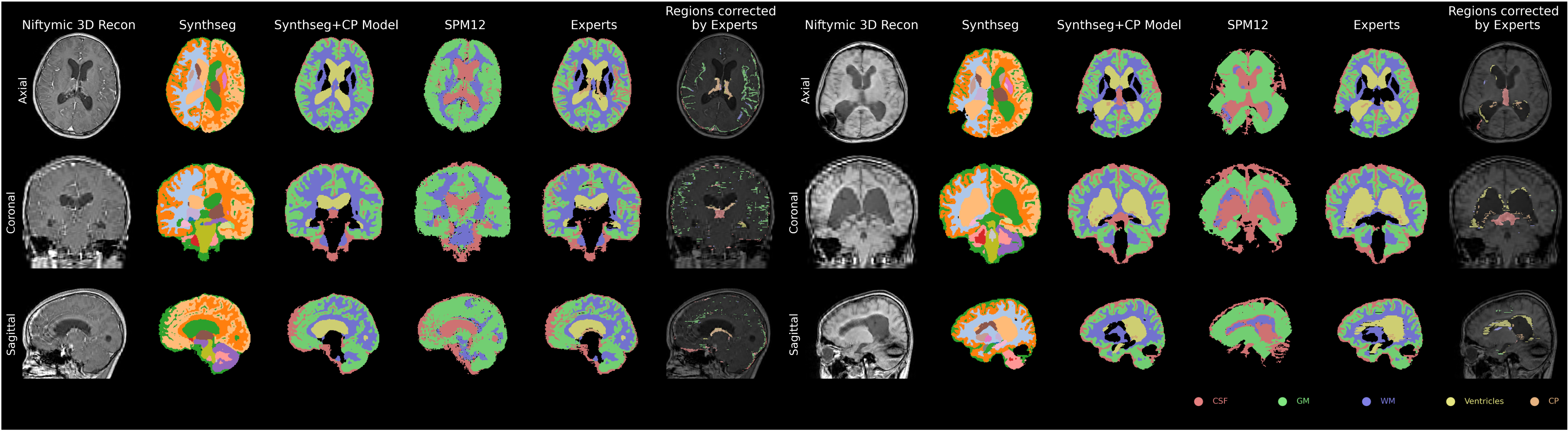}
    \caption{Comparison of segmentation methods. Results from three automated methods are compared against the physician-corrected gold standard in two pediatric hydrocephalus cases with differing severity: mild (left) versus severe (right) ventriculomegaly. The 'Regions corrected by Experts' panel displays the manual corrections based on the Synthseg+CP Model segmentation. }
	\label{fig:segmentation}
\end{figure*}
 
\paragraph{Demographics and Acquisition Parameters:}
Patients included in this study met the following criteria: 1) A clinical diagnosis of pediatric hydrocephalus; 2) An age range from neonate to adolescent (0–17 years); 3) Underwent a 3T MRI examination, with all scans performed on Philips MRI systems. Exclusion criteria were as follows: 1) Poor image quality that precluded effective super-resolution reconstruction; 2) Absence of an imaging report.

The dataset comprised 50 scans from 48 patients. Detailed demographic and MRI information is provided in Table \ref{tab:Demographic}. The age and sex distributions are shown in Figure \ref{fig:demographics}.a. Patients younger than 5 years constituted 36\% of the cohort. According to radiology reports, Figure \ref{fig:demographics}.b, ventricular volume was stable in most cases (30 patients). In terms of overall condition assessment, 31 patients were stable, 14 showed improvement, and 3 showed deterioration. 28 patients were with available surgical records, 19 underwent VP shunting, 6 had an endoscopic third ventriculostomy, and the remaining 3 underwent tumor resection. Ward round reports (i.e., clinical progress notes) from 29 cases documented common complications, including infection, shunt obstruction, and overdrainage.

\section{Method}
\subsection{Slice-to-Volume Super-Resolution Reconstruction}
\begin{figure}[h]
    \centering
    \includegraphics[width=0.7\linewidth]{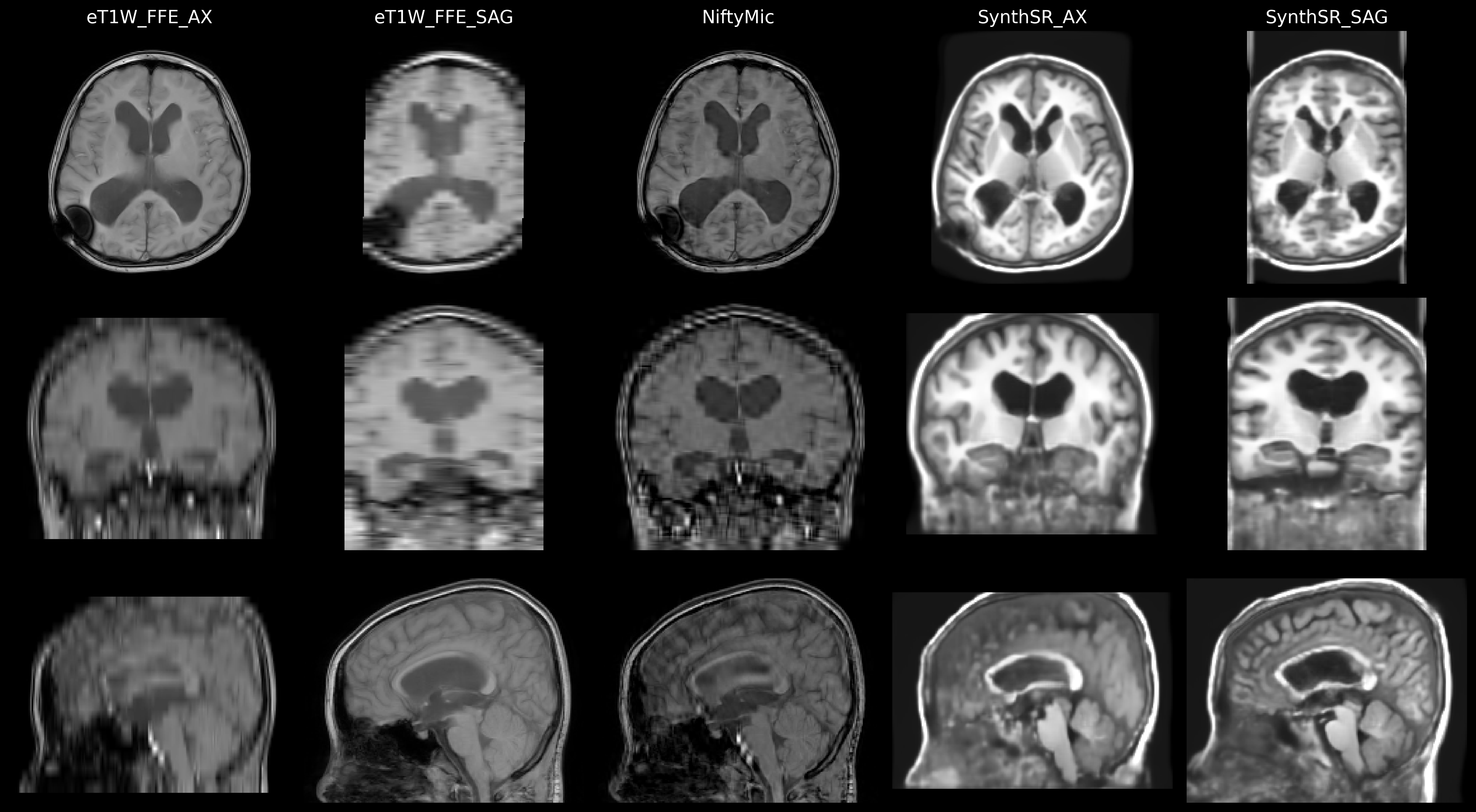}
    \caption{Visual comparison NiftyMIC versus SynthSR.}
	\label{fig:SynthSR}
\end{figure}

To overcome the limitations of anisotropy and low resolution in the clinical data, a slice-to-volume reconstruction (SVR) technique, NiftyMIC \citep{ebnerAutomatedFrameworkLocalization2020}, was employed. Each case was acquired using fast-scan sequence at both axial and sagittal planes, which had a large slice thickness of about 7mm. The SVR began with converting the de-identified DICOM images to the NifTI format using SimpleITK version 2.4.1. Then, the initial manual orientation correction was performed on the axial and sagittal images to correct orientation by Y.X, a PhD candidate with four years of experience in medical image processing. Finally, the open-source NiftyMIC toolkit   \citep{ebnerAutomatedFrameworkLocalization2020} was used as the baseline for SVR to generate the 1mm isotropic 3D volumes. The SynthSR \citep{iglesias2023synthsr} was also evaluated; despite yielding fewer artifacts, its output suffered from structural distortion and blurring, Figure \ref{fig:SynthSR}. Therefore, the experts selected the NiftyMIC reconstructions for manual annotation.

\subsection{Tissue  Segmentation}
Accurate segmentation of brain tissues and ventricles is crucial for the quantitative assessment of hydrocephalus. Based on the reconstructed high-resolution 3D images, thr-\\ee automated tools were used to obtain initial segmentations, including SynthSeg \citep{billotSynthSegSegmentationBrain2023}, SPM12 \citep{penny2011statistical}, and an in-house choroid plexus model \citep{liAssociationsChoroidPlexus2024}. Specifically, SynthSeg was used to generate segmentation labels for 61 brain regions. SPM12, which leverages Tissue Probability Maps and a unified segmentation model, was employed to generate probability maps for gray matter (GM), white matter (WM), and external  CSF. 
The CP model utilized a two-stage 3D U-Net, which first localized the lateral ventricle to define a region of interest, and then segmented the CP within that cropped region. This model was subsequently refined through a human-in-the-loop active learning strategy, where an expert iteratively corrected the poorest segmentations to augment the training dataset.\citep{liAssociationsChoroidPlexus2024}
The labels of initial segmentations were combined into GM, WM, external CSF, lateral ventricle, and CP regions and provided to a neurologist for manual correction. This segmentations could be mapped back to the original low-resolution images.

\begin{figure*}[htb]
	\centering
	\includegraphics[width=0.8\linewidth]{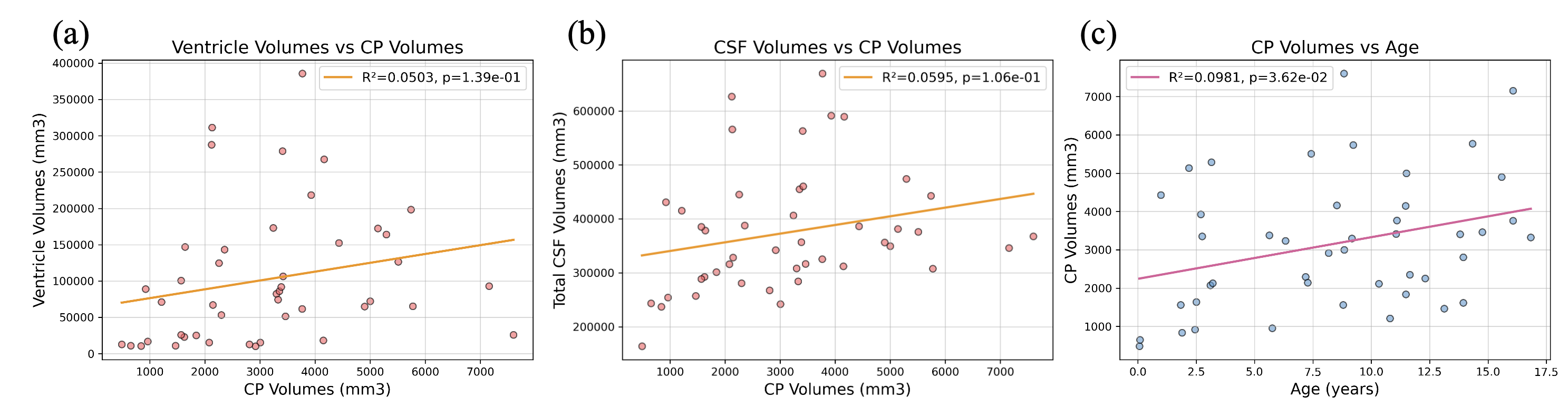}
	\caption{Linear Regression Analysis of Choroid Plexus Volume, Ventricle Volume, total CSF  Volume, and Age. The CSF Volume refers to TCV in the Figure.}
	\label{fig:correlation_analysis}
\end{figure*}

\subsection{RAG-based Structuring of Clinical Information}
A RAG framework was employed to establish a structured clinical report. A knowledge base was constructed by compiling 10 clinical guidelines and cutting-edge surgical papers related to hydrocephalus. This knowledge base was then vectorized using zhipu-embedding-v3 \citep{zhipu-embedding-v3} and stored in a FAISS index \citep{johnson2019billion}. This knowledge base was then used to augment a large language model, qwen3-235b-A22b  \citep{yang2025qwen3}, to automatically extract key structured information from unstructured surgical notes, ward round reports, and imaging reports. It extracted structured clinical information for each patient, including hydrocephalus etiology, imaging features, surgical details, and postoperative status.

Among the extracted data, this study specifically focused on three structured variables to investigate the role of the choroid plexus: etiology, ventricular change trend, and condition severity trend. The hydrocephalus etiology was determined from the clinical course from imaging reports, the ventricular trend was derived from serial imaging reports, and the severity trend was synthesized from a holistic assessment of imaging, ward, and surgical notes.

\begin{table}[htbp]
	\centering
	\caption{Comparison of Dice Coefficients (DSC) for brain tissue segmentation between automated methods and the expert gold standard.} 
	\resizebox{0.48\textwidth}{!}{%
	\begin{tabular}{lccccc}
	  \toprule
	  & \multicolumn{5}{c}{Tissue Type} \\ 
	  \cmidrule(lr){2-6} 
	  Method & External CSF & GM & WM & Ventricle & CP \\
	  \midrule
	  Synthseg vs. Expert & $0.9281 \pm 0.0341$ & $0.9740 \pm 0.0260$ & $0.9726 \pm 0.0528$ & $0.8663 \pm 0.2021$ & $0.4311 \pm 0.2360$ \\
	  \addlinespace 
	  SPM12 vs. Expert & $0.2790 \pm 0.0726$ & $0.5943 \pm 0.0780$ & $0.4557 \pm 0.2964$ & --- & --- \\
	  \bottomrule
	\end{tabular}
	}
	\label{tab:tissue_comparison} 
\end{table}

\section{Benchmarks}
This dataset is intended to provide a comprehensive benchmark platform for research in pediatric hydrocephalus. We have established the primary data:

\paragraph{Super-Resolution Reconstruction:} The dataset provides the original low-resolution orthogonal scans and the corresponding baseline reconstruction results from NiftyMIC. This allows researchers to develop novel super reolution algorithms and compare their performance against our established baseline.
\paragraph{Brain Tissue Segmentation: } Qualitative segmentation results from the three automated methods are shown in Figure \ref{fig:segmentation}.
The expert segmentations involved substantial manual corrections, particularly on the ventricles, CSF and the CP, Figure \ref{fig:segmentation}.
The Dice coefficients between these results and the gold  standard are shown in Table \ref{tab:tissue_comparison}. The results clearly indicate that SPM12 performs poorly on this task and that automated segmentation of the CP is particularly challenging. This highlights the need for specialized segmentation algorithms for hydrocephalus, a research direction this dataset aims to advance.

\section{ Association between Choroid Plexus and CSF Volume}
To evaluate the association between choroid plexus volume and CSF volume, and its clinical heterogeneity in pediatric hydrocephalus, a multi-level statistical analysis was performed. Correlation analyses were performed using choroid plexus volume (CPV) as the independent variable, with both ventricular volume (VV) and total CSF volume (TCV, the sum of ventricular and extraventricular CSF volumes) as dependent variables. 
All these volumes were quantified based on the gold standard of manual segmentation by a neurologist, with the final measurement derived from voxel summation.
This approach aims to more fully capture the role of the CP in the pathophysiology of hydrocephalus.
\begin{figure*}[h]
    \centering
	\includegraphics[width=0.85\linewidth]{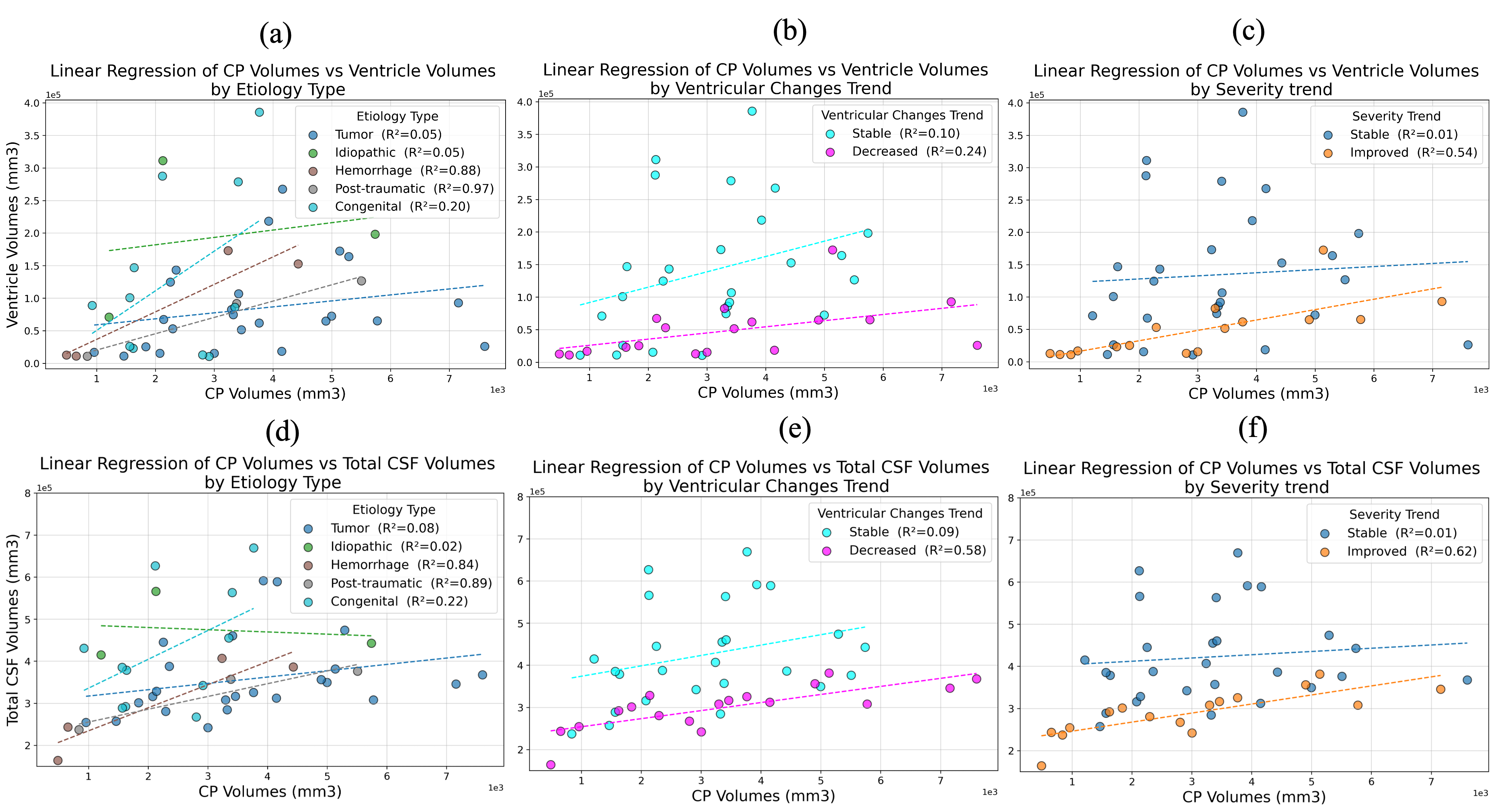}
    \caption{ Linear regression plots for the stratified analysis of the association between CPV and VV/TCV. Subgroups stratified by etiology (left), ventricular trend (middle), and clinical severity trend (right) were presented, respectively. The relationship between CPV and VV was shown in the top row, and the relationship between CPV and TCV was shown in the bottom row.}
	\label{fig:stratified_analysis}
\end{figure*}

\subsection{Baseline Correlation Analysis}
In the baseline correlation analysis, we assessed the relationships between age and CPV, between CPV and VV, and between CPV and TCV, using both linear regression and Spearman's correlation.
\subsection{Stratified Correlation} 
In addition, stratified correlation and covariance analysis was performed. Within each of the three clinical subgroups, including etiology (tumor, congenital, hemorrhage, post-traumatic and idiopathic), ventricular trend (stable and decreased), severity trend (stable and improved), the above correlation analyses were performed. Then ANCOVA \citep{kutner2005applied} was applied, using age as a covariate, to test for significant differences in these relationships among the subgroups.

\subsection{Prediction Model} 
Prediction Models were developed, in which Logistic Regression (LR) models were constructed using age, CPV, and either VV or TCV as input features to predict ventricular trend and severity trend. The performance of the proposed models was evaluated using 5-fold cross-validation.

\section{Results}
\subsection{Choroid Plexus Effects}

As shown in Figure \ref{fig:correlation_analysis} and Table \ref{tab:correlation_analysis}, Spearman's correlation analysis revealed a significant and positive correlation between CPV and VV ($\rho$=0.42, p = 0.0045), CPV and TCV($\rho$=0.37, p = 0.013), indicating a monotonic increase between the CPV and VV or TCV in hydrocephalus. However, the linear regression model did not show a significant linear relationship (R² = 0.05, p = 0.139). Concurrently, a significant and positive correlation was found between CPV and age (Spearman's $\rho$ = 0.33, p = 0.027). Although the linear regression between the CPV and age was statistically significant (p = 0.036), the low coefficient of determination (R² = 0.098) indicated limited explanatory power given the sample size.
\begin{table}[htbp]
	\centering
	  \caption{Correlation and Regression Analysis of Choroid Plexus Volume, Ventricle Volume, total CSF Volume, and Age.}
        \begin{threeparttable}
	  \label{tab:correlation_analysis}
	  \resizebox{0.48\textwidth}{!}{%
	  \begin{tabular}{lllr}
		\toprule
		Variable Pair & Analysis Method & Coefficient & p-value \\
		\midrule
		\multirow{2}{*}{CPV vs. VV} & Spearman's Correlation & $\rho = 0.420$ & \textbf{0.005} \\
																  & Linear Regression      & $R^2 = 0.050$  & 0.139 \\
		\addlinespace 
		\multirow{2}{*}{CPV vs. TCV}       & Spearman's Correlation & $\rho = 0.369$ & \textbf{0.013} \\
																  & Linear Regression      & $R^2 = 0.0595$ & 0.106 \\
		\addlinespace
		\multirow{2}{*}{Age vs. CPV}              & Spearman's Correlation & $\rho = 0.330$ & \textbf{0.027} \\
																  & Linear Regression      & $R^2 = 0.098$  & \textbf{0.036} \\
		\bottomrule
	  \end{tabular}
	  }
	\end{threeparttable}
	\label{tab:correlation_analysis}
\end{table}

\begin{table*}[htbp]
	\centering
	\caption{Stratified Analysis of the Association between Choroid Plexus (CP) Volume and CSF Compartment Volumes. This table details the results of correlation (Spearman's $\rho$) and linear regression (R²) analyses for the relationship between CPV VV/TCV. The analyses are stratified by clinical subgroups based on etiology, radiological ventricular trend, and clinical Severity trend. Key statistics, including sample size (N) and p-values, are presented for each subgroup.}
	\label{tab:clinical_analysis}
	\resizebox{0.8\textwidth}{!}{%
	\begin{tabular}{lllcccccccccc}
	\toprule
	\multirow{2}{*}{\textbf{Group}} & \multirow{2}{*}{\textbf{Clinical Outcome}} & \multirow{2}{*}{\textbf{Samples}} & \multicolumn{5}{c}{\textbf{Ventricle Volume}} & \multicolumn{5}{c}{\textbf{Total CSF Volume}} \\
	\cmidrule(lr){4-8} \cmidrule(lr){9-13}
	& & & \multicolumn{3}{c}{\textbf{Linear Regression}} & \multicolumn{2}{c}{\textbf{spearman}} & \multicolumn{3}{c}{\textbf{Linear Regression}} & \multicolumn{2}{c}{\textbf{spearman}} \\
	\cmidrule(lr){4-6} \cmidrule(lr){7-8} \cmidrule(lr){9-11} \cmidrule(lr){12-13}
	& & & \textbf{$R^2$} & \textbf{F-statistic} & \textbf{p-value} & \textbf{Coefficient} & \textbf{p-value} & \textbf{$R^2$} & \textbf{F-statistic} & \textbf{p-value} & \textbf{Coefficient} & \textbf{p-value} \\
	\midrule
	\multirow{5}{*}{Etiology} 
	& Tumor & 24 & 0.0538 & 1.25 & 0.2756 & 0.4322 & \textbf{0.0349} & 0.0754 & 1.7951 & 0.194 & 0.4887 & \textbf{0.0154} \\
	& Idiopathic & 3 & 0.0509 & 0.0536 & 0.8551 & 0.5 & 0.6667 & 0.025 & 0.0256 & 0.899 & 0.5 & 0.6667 \\
	& Hemorrhage & 4 & 0.8815 & 14.8809 & 0.0611 & 0.6 & 0.4 & 0.8438 & 10.8001 & 0.0814 & 0.8 & 0.2 \\
	& Post-traumatic & 3 & 0.9694 & 31.629 & 0.112 & 1 & 0 & 0.8857 & 7.7516 & 0.2195 & 1 & 0 \\
	& Congenital & 11 & 0.1994 & 2.2421 & 0.1685 & 0.2364 & 0.4841 & 0.2204 & 2.5442 & 0.1452 & 0.4364 & 0.1797 \\
	\midrule
	\multirow{2}{*}{\begin{tabular}[c]{@{}l@{}}Ventricular \\ Changes Trend\end{tabular}} & Stable & 26 & 0.103 & 2.7562 & 0.1099 & 0.4619 & \textbf{0.0175} & 0.0838 & 2.2535 & 0.1464 & 0.386 & 0.0515 \\
	& Decreased & 18 & 0.2354 & 4.9256 & \textbf{0.0413} & 0.6367 & \textbf{0.0045} & 0.5778 & 21.8956 & \textbf{0.0003} & 0.7771 & \textbf{0.0001} \\
	\midrule
	\multirow{2}{*}{\begin{tabular}[c]{@{}l@{}}Severity \\ Trend\end{tabular}} & Stable & 28 & 0.0054 & 0.1399 & 0.7114 & 0.2649 & 0.1731 & 0.0115 & 0.3029 & 0.5868 & 0.2178 & 0.2654 \\
	& Improved & 16 & 0.5397 & 16.4127 & \textbf{0.0012} & 0.8647 & \textbf{1e-5} & 0.6183 & 22.6748 & \textbf{0.0003} & 0.8471 & \textbf{3e-4} \\
	\bottomrule
	\end{tabular}%
	}
	\begin{tablenotes}
		\item[*] Statistically significant p-values ($p < 0.05$) are shown in bold.
	\end{tablenotes}
	\label{tab:Stratified_Analysis} 

\end{table*}

\subsection{Stratified Analysis of CP-CSF Association}
As shown in Figure \ref{fig:stratified_analysis} and Table \ref{tab:Stratified_Analysis}, the stratified analysis revealed association patterns that were highly dependent on the clinical subgroups. In both the "Decreased" ventricular (N=18) and the "Improved" severity trend group (N=16), CPV exhibited a significant positive correlation with VV (p = 0.0012). In contrast, the "Stable" ventricular trend group (N=26) showed only a moderate positive correlation, while the correlation in the "Stable" severity trend group (N=28) was not significant. 

The results in the "Decreased" ventricular volume group showed high linear correlation, with R² increasing from 0.23 to 0.57, the Spearman coefficient rising from 0.63 to 0.77, and all p-values demonstrating greater statistical significance when the dependent variable was switched to TCV. This finding suggests that the correlation between CPV and VV/TCV may be a key indicator of improvement, signifying ventricular reduction and clinical betterment.

In the stratification by etiology, a significant and moderate positive correlation was observed between CPV and VV  ($\rho$ = 0.432, p = 0.035) in the tumor subgroup (N=24). This correlation was enhanced when using TCV as the dependent variable, which had an increase in R² by 0.21 and a reduction in the p-value from 0.0349 to 0.0154 in linear regression.

\subsection{Multi-Group Comparision} 
As shown in Table \ref{tab:ancova_analysis_final}, after controlling for age as a covariate, ANCOVA showed significant differences in volume (for both VV and TCV) among the clinical subgroups ($p <0.05$). Specifically, the "Stable" groups had significantly larger volumes than the "Improved"/"Decreased" groups, which is consistent with the scatter distribution in the regression plots, Figure \ref{fig:stratified_analysis}. This may explain the weaker correlation in the "Stable" groups, as the exceptionally large volumes could attenuate the linear relationship with CPV. Importantly, the interaction effect between CPV and the clinical subgroups was not significant, indicating that the influence of CPV on ventricular/total CSF volume is consistent across these subgroups.

 The model's explanatory power increased when TCV replaced VV as the dependent variable (from 0.327 to 0.403 in the ventricular trend group). More critically, after controlling for subgroup and age, the independent effect of CPV as a covariate on TCV was more significant (F=6.868, p = 0.012). This result reveals that TCV more robustly reflects the central role of CPV in the pathophysiology of hydrocephalus.

\subsection{LR Model for Clinical Prediction}
\begin{figure}[h]
    \centering
    \includegraphics[width=1\linewidth]{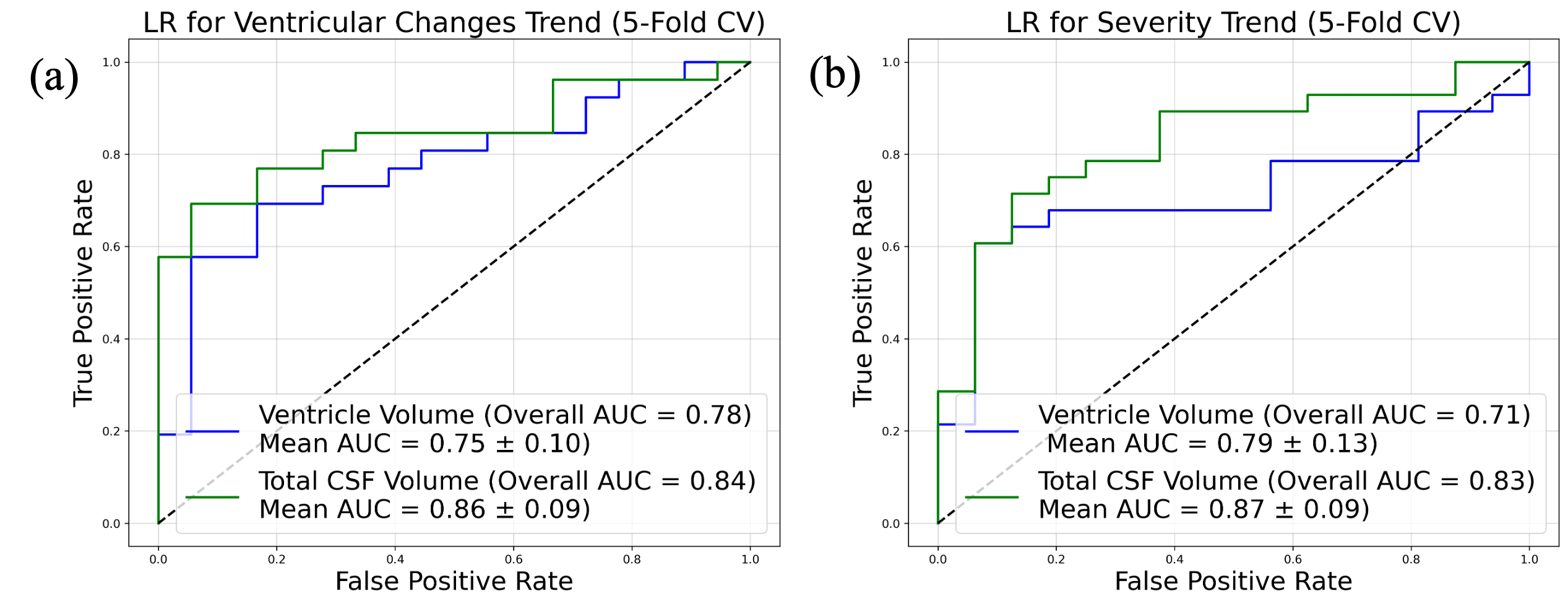}
    \caption{ROC curves for predicting (a) ventricular trend and (b) clinical severity trend. The curves compare models using either VV or TCV, adjusted for age and CPV.}
	\label{fig:logistic_regression}
\end{figure}

The clinical prediction models built on LR showed substantially improved performance when both CPV and TCV were included as features, Figure \ref{fig:logistic_regression}. In predicting the ventricular trend, the mean Area Under the Curve (AUC) from five-fold cross-validation increased from 0.75 (CPV and VV) to 0.86 ± 0.09. For predicting the severity trend, the mean AUC increased from 0.79 to 0.87 ± 0.09.
These results indicate that the model integrating the CP-TCV association achieves excellent predictive performance.

\begin{table}[htbp]
	\centering
	\caption{Analysis of Covariance (ANCOVA) for Ventricle and Total CSF Volumes Stratified by Ventricular Change and Severity Trend, with Choroid Plexus Volume as a Covariate.}
        \resizebox{0.5\textwidth}{!}{%
	\begin{tabular}{lllccc}
	\toprule
	\textbf{Group} & \textbf{Source of Variation} & \textbf{Analysis} & \textbf{Degree of Freedom} & \textbf{F-value} & \textbf{p-value} \\
	\midrule
	\multirow{10}{*}{\begin{tabular}[c]{@{}l@{}}Ventricular \\ Trend\end{tabular}} 
	& \multirow{5}{*}{\textbf{Ventricle Volume}} 
      & Between-group difference & 1 & 15.5824 & \textbf{0.0003} \\
	& & CPV (Covariate Effect) & 1 & 4.346 & \textbf{0.0434} \\
	& & Interaction Effect & 1 & 0.9139 & 0.345 \\
	& & Residual & 41 & & \\
	& & Explanatory Power & 0.327 & & \\
	\cmidrule(l){2-6} 
	& \multirow{5}{*}{\textbf{Total CSF Volume}} 
      & Between-group difference & 1 & 22.7225 & \textbf{2e-4} \\
	& & CPV (Covariate Effect) & 1 & 6.8683 & \textbf{0.0123} \\
	& & Interaction Effect & 1 & 3.5494 & 0.067 \\
	& & Residual & 41 & & \\
	& & Explanatory Power & 0.403 & & \\
	\midrule
	\multirow{10}{*}{\begin{tabular}[c]{@{}l@{}}Severity \\ Trend\end{tabular}} 
	& \multirow{5}{*}{\textbf{Ventricle Volume}} 
      & Between-group difference & 1 & 9.7356 & \textbf{0.0033} \\
	& & CPV (Covariate Effect) & 1 & 1.7934 & 0.1879 \\
	& & Interaction Effect & 1 & 0.5417 & 0.466 \\
	& & Residual & 41 & & \\
	& & Explanatory Power & 0.243 & & \\
    \cmidrule(l){2-6} 
	& \multirow{5}{*}{\textbf{Total CSF Volume}} 
      & Between-group difference & 1 & 19.6529 & \textbf{0.0001} \\
	& & CPV (Covariate Effect) & 1 & 2.9421 & 0.0938 \\
	& & Interaction Effect & 1 & 3.16 & 0.083 \\
	& & Residual & 41 & & \\
	& & Explanatory Power & 0.382 & & \\
	\bottomrule
	\end{tabular}
    }
	\label{tab:ancova_analysis_final}
\end{table}

\subsection{Summary of Findings}

The choroid plexus and CSF coupling may provide a novel biomarker of pediatric hydrocephalus. The strong linear relationship between the volumes of the CP and the CSF is not constant among groups, and the correlation is associated to clinical improvement (ventricular reduction, symptomatic relief). This suggests that the CSF volume can be coupled with its production source, the choroid plexus.

Total CSF volume provided a superior explanatory variable in ANCOVA and LR result . Compared to ventricular volume alone, total CSF volume accurately and robustly reveals the intrinsic association between CP volume and the entire CSF space, making it a superior indicator.

The association between choroid plexus volume and total CSF volume may provide a novel indicator of clinical outcome. The LR model integrating CPV, TCV, and age demonstrated high predictive accuracy (AUC = 0.87), confirming the significant potential of the CP-CSF association as a clinical biomarker for assessing hydrocephalus.

\section{Discussion}
The present work shows that the association between choroid plexus (CP) volume and total CSF volume (TCV) is a potential biomarker of clinical status. A strong CP-CSF coupling was significantly correlated with clinical improvement.

In contrast, the monotonic, non-linear association observed in clinically stable patients suggests more complex pathophysiological mechanisms that warrant deeper exploration into their underlying causes. The inclusion of precise CP segmentations, a feature rarely available in public datasets, was critical to this analysis.

Furthermore, total CSF volume can provide a more robust metric than ventricular volume alone. This is physiologically plausible because TCV encapsulates the global state of intracranial fluid balance, not merely ventricular enlargement. The superior performance of total CSF volume in our predictive models underscores its value for assessing clinical status and its potential for tracking postoperative outcomes.

This study has several limitations. The single-center data and single expert manual correction may limit model generalizability. Methodologically, limitations include artifacts in some super-resolution reconstructions and constraints on clinical data extraction due to incomplete records and potential LLM  'hallucinations'. The weaker volumetric correlations in "stable" patients likely reflect the clinical heterogeneity of this cohort.

\section{Conclusion}
In this work, a comprehensive dataset for pediatric hydrocephalus were provided, including high-resolution reconstructed MRI, manually labeled segmentations, and RAG-extracted clinical features. This resource may provide critical data and benchmarks for hydrocephalus model developments. Based on this dataset, choroid plexus-CSF coupling was presented as a novel biomarker of clinical evaluation, and total CSF volume was shown to be an effective indicator compared to the conventional ventricle volume. Besides age, the choroid plexus and total CSF volume demonstrated accurate prediction of clinical symptoms using LR model, which is readily to be used in hydrocephalus studies.


\acks{This work is supported by the National Key R\&D Program of China (2023YFE0118900) and the MOE Frontier Science Center for Brain Science \& Brain-Machine Integration, Zhejiang University. }

%
\ethics{The work follows appropriate ethical standards in conducting research and writing the manuscript, following all applicable laws and regulations regarding treatment of animals or human subjects.}

\coi{The authors declare that they have no conflicts of interest.}
\data{The HyKid dataset is publicly available on an open-source platform to support research in the field.  The dataset is hosted at https://www.synapse.org/Synapse:syn68544889. To obtain the data, users are required to complete a short online application form and accept the terms of the Data Use Agreement.}
\bibliography{Hydrocephalus}

\begin{thebibliography}{33}
\providecommand{\natexlab}[1]{#1}
\providecommand{\url}[1]{\texttt{#1}}
\expandafter\ifx\csname urlstyle\endcsname\relax
  \providecommand{\doi}[1]{doi: #1}\else
  \providecommand{\doi}{doi: \begingroup \urlstyle{rm}\Url}\fi

\bibitem[Akinci~D’Antonoli et~al.(2023)Akinci~D’Antonoli, Todea, Leu, Datta, Stieltjes, Pruefer, and Wasserthal]{akinci2023development}
Tugba Akinci~D’Antonoli, Ramona-Alexandra Todea, Nora Leu, Alexandre~N Datta, Bram Stieltjes, Friederike Pruefer, and Jakob Wasserthal.
\newblock Development and evaluation of deep learning models for automated estimation of myelin maturation using pediatric brain mri scans.
\newblock \emph{Radiology: Artificial Intelligence}, 5\penalty0 (5):\penalty0 e220292, 2023.

\bibitem[Billot et~al.(2023)Billot, Greve, Puonti, Thielscher, Van~Leemput, Fischl, Dalca, and Iglesias]{billotSynthSegSegmentationBrain2023}
Benjamin Billot, Douglas~N. Greve, Oula Puonti, Axel Thielscher, Koen Van~Leemput, Bruce Fischl, Adrian~V. Dalca, and Juan~Eugenio Iglesias.
\newblock {SynthSeg}: {Segmentation} of brain {MRI} scans of any contrast and resolution without retraining.
\newblock \emph{Medical Image Analysis}, 86:\penalty0 102789, May 2023.
\newblock ISSN 1361-8415.
\newblock \doi{10.1016/j.media.2023.102789}.
\newblock URL \url{https://www.sciencedirect.com/science/article/pii/S1361841523000506}.

\bibitem[Damkier et~al.(2013)Damkier, Brown, and Praetorius]{damkierCerebrospinalFluidSecretion2013}
Helle~H Damkier, Peter~D Brown, and Jeppe Praetorius.
\newblock Cerebrospinal fluid secretion by the choroid plexus.
\newblock \emph{Physiological reviews}, 93\penalty0 (4):\penalty0 1847--1892, 2013.

\bibitem[Del~Bigio(2010)]{delbigioNeuropathologyStructuralChanges2010}
Marc~R. Del~Bigio.
\newblock Neuropathology and structural changes in hydrocephalus.
\newblock \emph{Developmental Disabilities Research Reviews}, 16\penalty0 (1):\penalty0 16--22, 2010.
\newblock ISSN 1940-5529.
\newblock \doi{10.1002/ddrr.94}.
\newblock URL \url{https://onlinelibrary.wiley.com/doi/abs/10.1002/ddrr.94}.
\newblock \_eprint: https://onlinelibrary.wiley.com/doi/pdf/10.1002/ddrr.94.

\bibitem[Ebner et~al.(2020)Ebner, Wang, Li, Aertsen, Patel, Aughwane, Melbourne, Doel, Dymarkowski, De~Coppi, et~al.]{ebnerAutomatedFrameworkLocalization2020}
Michael Ebner, Guotai Wang, Wenqi Li, Michael Aertsen, Premal~A Patel, Rosalind Aughwane, Andrew Melbourne, Tom Doel, Steven Dymarkowski, Paolo De~Coppi, et~al.
\newblock An automated framework for localization, segmentation and super-resolution reconstruction of fetal brain mri.
\newblock \emph{NeuroImage}, 206:\penalty0 116324, 2020.

\bibitem[Engelhardt et~al.(2015)Engelhardt, Inder, Alexopoulos, Dierker, Hill, Van~Essen, and Neil]{engelhardt2015regional}
Erin Engelhardt, Terrie~E Inder, Dimitrios Alexopoulos, Donna~L Dierker, Jason Hill, David Van~Essen, and Jeffrey~J Neil.
\newblock Regional impairments of cortical folding in premature infants.
\newblock \emph{Annals of neurology}, 77\penalty0 (1):\penalty0 154--162, 2015.

\bibitem[Evans(1942)]{evans1942encephalographic}
William~A Evans.
\newblock An encephalographic ratio for estimating ventricular enlargement and cerebral atrophy.
\newblock \emph{Archives of Neurology \& Psychiatry}, 47\penalty0 (6):\penalty0 931--937, 1942.

\bibitem[Flannery and Mitchell(2014)]{flanneryPediatricHydrocephalusSystematic2014}
Ann~Marie Flannery and Laura Mitchell.
\newblock Pediatric hydrocephalus: systematic literature review and evidence-based guidelines. {Part} 1: {Introduction} and methodology.
\newblock November 2014.
\newblock \doi{10.3171/2014.7.PEDS14321}.
\newblock URL \url{https://thejns.org/pediatrics/view/journals/j-neurosurg-pediatr/14/Supplement_1/article-p3.xml}.
\newblock Section: Journal of Neurosurgery: Pediatrics.

\bibitem[Fletcher et~al.(1992)Fletcher, Bohan, Brandt, Brookshire, Beaver, Francis, Davidson, Thompson, and Miner]{fletcher1992cerebral}
Jack~M Fletcher, Timothy~P Bohan, Michael~E Brandt, Bonnie~L Brookshire, Stephen~R Beaver, David~J Francis, Kevin~C Davidson, Nora~M Thompson, and Michael~E Miner.
\newblock Cerebral white matter and cognition in hydrocephalic children.
\newblock \emph{Archives of neurology}, 49\penalty0 (8):\penalty0 818--824, 1992.

\bibitem[Garegnani et~al.(2020)Garegnani, Franco, Ciapponi, Garrote, Vietto, and Medina]{garegnaniVentriculoperitonealShuntingDevices2020}
Luis Garegnani, Juan~VA Franco, Agustín Ciapponi, Virginia Garrote, Valeria Vietto, and Santiago Adalberto~Portillo Medina.
\newblock Ventriculo‐peritoneal shunting devices for hydrocephalus.
\newblock \emph{Cochrane Database of Systematic Reviews}, \penalty0 (6), 2020.
\newblock ISSN 1465-1858.
\newblock \doi{10.1002/14651858.CD012726.pub2}.
\newblock URL \url{https://www.cochranelibrary.com/cdsr/doi/10.1002/14651858.CD012726.pub2/full}.
\newblock Publisher: John Wiley \& Sons, Ltd.

\bibitem[Ha et~al.(2020)Ha, Baek, Ryu, Choi, Moon, Park, and Kim]{haOneMinuteUltrafastBrain2020}
Ji~Young Ha, Hye~Jin Baek, Kyeong~Hwa Ryu, Bo~Hwa Choi, Jin~Il Moon, Sung~Eun Park, and Tae~Byeong Kim.
\newblock One-{Minute} {Ultrafast} {Brain} {MRI} {With} {Full} {Basic} {Sequences}: {Can} {It} {Be} a {Promising} {Way} {Forward} for {Pediatric} {Neuroimaging}?
\newblock \emph{American Journal of Roentgenology}, 215\penalty0 (1):\penalty0 198--205, July 2020.
\newblock ISSN 0361-803X, 1546-3141.
\newblock \doi{10.2214/AJR.19.22378}.
\newblock URL \url{https://www.ajronline.org/doi/10.2214/AJR.19.22378}.

\bibitem[Hunt et~al.(2003)Hunt, Warfield, Wang, Kean, Volpe, and Inder]{hunt2003assessment}
RW~Hunt, SK~Warfield, H~Wang, M~Kean, JJ~Volpe, and Terrie~E Inder.
\newblock Assessment of the impact of the removal of cerebrospinal fluid on cerebral tissue volumes by advanced volumetric 3d-mri in posthaemorrhagic hydrocephalus in a premature infant.
\newblock \emph{Journal of Neurology, Neurosurgery \& Psychiatry}, 74\penalty0 (5):\penalty0 658--660, 2003.

\bibitem[Iglesias et~al.(2023)Iglesias, Billot, Balbastre, Magdamo, Arnold, Das, Edlow, Alexander, Golland, and Fischl]{iglesias2023synthsr}
Juan~E Iglesias, Benjamin Billot, Ya{\"e}l Balbastre, Colin Magdamo, Steven~E Arnold, Sudeshna Das, Brian~L Edlow, Daniel~C Alexander, Polina Golland, and Bruce Fischl.
\newblock Synthsr: A public ai tool to turn heterogeneous clinical brain scans into high-resolution t1-weighted images for 3d morphometry.
\newblock \emph{Science advances}, 9\penalty0 (5):\penalty0 eadd3607, 2023.

\bibitem[Johnson et~al.(2019)Johnson, Douze, and J{\'e}gou]{johnson2019billion}
Jeff Johnson, Matthijs Douze, and Herv{\'e} J{\'e}gou.
\newblock Billion-scale similarity search with gpus.
\newblock \emph{IEEE Transactions on Big Data}, 7\penalty0 (3):\penalty0 535--547, 2019.

\bibitem[Kahle et~al.(2016)Kahle, Kulkarni, Limbrick, and Warf]{kahleHydrocephalusChildren2016a}
Kristopher~T Kahle, Abhaya~V Kulkarni, David~D Limbrick, and Benjamin~C Warf.
\newblock Hydrocephalus in children.
\newblock \emph{The Lancet}, 387\penalty0 (10020):\penalty0 788--799, February 2016.
\newblock ISSN 0140-6736.
\newblock \doi{10.1016/S0140-6736(15)60694-8}.
\newblock URL \url{https://www.sciencedirect.com/science/article/pii/S0140673615606948}.

\bibitem[Kestle et~al.(2011)Kestle, Riva-Cambrin, Wellons, Kulkarni, Whitehead, Walker, Oakes, Drake, Luerssen, Simon, and Holubkov]{kestleStandardizedProtocolReduce2011}
John R.~W. Kestle, Jay Riva-Cambrin, John~C. Wellons, Abhaya~V. Kulkarni, William~E. Whitehead, Marion~L. Walker, W.~Jerry Oakes, James~M. Drake, Thomas~G. Luerssen, Tamara~D. Simon, and Richard Holubkov.
\newblock A standardized protocol to reduce cerebrospinal fluid shunt infection: {The} {Hydrocephalus} {Clinical} {Research} {Network} {Quality} {Improvement} {Initiative}.
\newblock \emph{Journal of neurosurgery. Pediatrics}, 8\penalty0 (1):\penalty0 22--29, July 2011.
\newblock ISSN 1933-0707.
\newblock \doi{10.3171/2011.4.PEDS10551}.
\newblock URL \url{https://www.ncbi.nlm.nih.gov/pmc/articles/PMC3153415/}.

\bibitem[Kulkarni et~al.(2013)Kulkarni, Riva-Cambrin, Butler, Browd, Drake, Holubkov, Kestle, Limbrick, Simon, Tamber, Wellons~III, and Whitehead]{kulkarniOutcomesCSFShunting2013}
A.V. Kulkarni, J.~Riva-Cambrin, J.~Butler, S.R. Browd, J.M. Drake, R.~Holubkov, J.R.W. Kestle, D.D. Limbrick, T.D. Simon, M.S. Tamber, J.C. Wellons~III, and W.E. Whitehead.
\newblock Outcomes of {CSF} shunting in children: {Comparison} of {Hydrocephalus} {Clinical} {Research} {Network} cohort with historical controls.
\newblock \emph{Journal of Neurosurgery: Pediatrics}, 12\penalty0 (4):\penalty0 334--338, 2013.
\newblock \doi{10.3171/2013.7.PEDS12637}.

\bibitem[Kutner(2005)]{kutner2005applied}
Michael~H Kutner.
\newblock Applied linear statistical models.
\newblock 2005.

\bibitem[Laurence and Coates(1962)]{laurenceNaturalHistoryHydrocephalus1962}
K.~M. Laurence and Stephen Coates.
\newblock The {Natural} {History} of {Hydrocephalus}.
\newblock \emph{Archives of Disease in Childhood}, 37\penalty0 (194):\penalty0 345--362, August 1962.
\newblock ISSN 0003-9888.
\newblock \doi{10.1136/adc.37.194.345}.
\newblock URL \url{https://www.ncbi.nlm.nih.gov/pmc/articles/PMC2012878/}.

\bibitem[Lewis et~al.(2020)Lewis, Perez, Piktus, Petroni, Karpukhin, Goyal, K{\"u}ttler, Lewis, Yih, Rockt{\"a}schel, et~al.]{lewis2020retrieval}
Patrick Lewis, Ethan Perez, Aleksandra Piktus, Fabio Petroni, Vladimir Karpukhin, Naman Goyal, Heinrich K{\"u}ttler, Mike Lewis, Wen-tau Yih, Tim Rockt{\"a}schel, et~al.
\newblock Retrieval-augmented generation for knowledge-intensive nlp tasks.
\newblock \emph{Advances in neural information processing systems}, 33:\penalty0 9459--9474, 2020.

\bibitem[Li et~al.(2024)Li, Hu, Xu, Feng, Meyer, Dai, Zhao, and {for the Alzheimer’s Disease Neuroimaging Initiative}]{liAssociationsChoroidPlexus2024}
Jiaxin Li, Yueqin Hu, Yunzhi Xu, Xue Feng, Craig~H. Meyer, Weiying Dai, Li~Zhao, and {for the Alzheimer’s Disease Neuroimaging Initiative}.
\newblock Associations between the choroid plexus and tau in {Alzheimer}’s disease using an active learning segmentation pipeline.
\newblock \emph{Fluids and Barriers of the CNS}, 21\penalty0 (1):\penalty0 56, July 2024.
\newblock ISSN 2045-8118.
\newblock \doi{10.1186/s12987-024-00554-4}.
\newblock URL \url{https://doi.org/10.1186/s12987-024-00554-4}.

\bibitem[Lindquist et~al.(2006)Lindquist, Carlsson, Persson, and Uvebrant]{lindquist2006behavioural}
Barbro Lindquist, G{\"o}ran Carlsson, Eva-Karin Persson, and Paul Uvebrant.
\newblock Behavioural problems and autism in children with hydrocephalus: a population-based study.
\newblock \emph{European child \& adolescent psychiatry}, 15:\penalty0 214--219, 2006.

\bibitem[McAllister(2012)]{mcallisterPathophysiologyCongenitalNeonatal2012}
James~P. McAllister.
\newblock Pathophysiology of congenital and neonatal hydrocephalus.
\newblock \emph{Seminars in Fetal and Neonatal Medicine}, 17\penalty0 (5):\penalty0 285--294, October 2012.
\newblock ISSN 1744-165X.
\newblock \doi{10.1016/j.siny.2012.06.004}.
\newblock URL \url{https://www.sciencedirect.com/science/article/pii/S1744165X12000790}.

\bibitem[Melot et~al.(2016)Melot, Labarre, Vanhulle, Rondeau, Brasseur, Gilard, Castel, Marret, and Proust]{melotNeurodevelopmentalLongtermOutcome2016}
A.~Melot, A.~Labarre, C.~Vanhulle, S.~Rondeau, M.~Brasseur, V.~Gilard, H.~Castel, S.~Marret, and F.~Proust.
\newblock Neurodevelopmental long-term outcome in children with hydrocephalus requiring neonatal surgical treatment.
\newblock \emph{Neurochirurgie}, 62\penalty0 (2):\penalty0 94--99, April 2016.
\newblock ISSN 0028-3770.
\newblock \doi{10.1016/j.neuchi.2015.10.009}.
\newblock URL \url{https://www.sciencedirect.com/science/article/pii/S0028377015003008}.

\bibitem[Patwardhan and Nanda(2005)]{patwardhanImplantedVentricularShunts2005}
Ravish~V. Patwardhan and Anil Nanda.
\newblock Implanted {Ventricular} {Shunts} in the {United} {States}: {The} {Billion}-dollar-a-year {Cost} of {Hydrocephalus} {Treatment}.
\newblock \emph{Neurosurgery}, 56\penalty0 (1):\penalty0 139--145, January 2005.
\newblock ISSN 0148-396X, 1524-4040.
\newblock \doi{10.1227/01.NEU.0000146206.40375.41}.
\newblock URL \url{http://journals.lww.com/10.1227/01.NEU.0000146206.40375.41}.

\bibitem[Penny et~al.(2011)Penny, Friston, Ashburner, Kiebel, and Nichols]{penny2011statistical}
William~D Penny, Karl~J Friston, John~T Ashburner, Stefan~J Kiebel, and Thomas~E Nichols.
\newblock \emph{Statistical parametric mapping: the analysis of functional brain images}.
\newblock Elsevier, 2011.

\bibitem[Rekate(2009)]{rekateContemporaryDefinitionClassification2009}
Harold~L. Rekate.
\newblock A {Contemporary} {Definition} and {Classification} of {Hydrocephalus}.
\newblock \emph{Seminars in Pediatric Neurology}, 16\penalty0 (1):\penalty0 9--15, March 2009.
\newblock ISSN 1071-9091.
\newblock \doi{10.1016/j.spen.2009.01.002}.
\newblock URL \url{https://www.sciencedirect.com/science/article/pii/S1071909109000059}.

\bibitem[Richards et~al.(2009)Richards, Seeley, and Pickard]{richards2009efficacy}
Hugh~K Richards, Helen~M Seeley, and John~D Pickard.
\newblock Efficacy of antibiotic-impregnated shunt catheters in reducing shunt infection: data from the united kingdom shunt registry.
\newblock \emph{Journal of Neurosurgery: Pediatrics}, 4\penalty0 (4):\penalty0 389--393, 2009.

\bibitem[Ros et~al.(2021)Ros, Iglesias, Linares, Cerro, Casado, and Arráez]{rosShuntOverdrainageReappraisal2021}
Bienvenido Ros, Sara Iglesias, Jorge Linares, Laura Cerro, Julia Casado, and Miguel~Angel Arráez.
\newblock Shunt {Overdrainage}: {Reappraisal} of the {Syndrome} and {Proposal} for an {Integrative} {Model}.
\newblock \emph{Journal of Clinical Medicine}, 10\penalty0 (16):\penalty0 3620, January 2021.
\newblock ISSN 2077-0383.
\newblock \doi{10.3390/jcm10163620}.
\newblock URL \url{https://www.mdpi.com/2077-0383/10/16/3620}.
\newblock Number: 16 Publisher: Multidisciplinary Digital Publishing Institute.

\bibitem[Sedano et~al.(2023)Sedano, Kronman, Whitlock, Zhou, Coffin, Hauptman, Heller, Mangano, Pollack, Schaffzin, et~al.]{sedanoAssociationsStandardCare2023a}
Sabrina Sedano, Matthew~P Kronman, Kathryn~B Whitlock, Chuan Zhou, Susan~E Coffin, Jason~S Hauptman, Evan Heller, Francesco~T Mangano, Ian~F Pollack, Joshua~K Schaffzin, et~al.
\newblock Associations of standard care, intrathecal antibiotics, and antibiotic-impregnated catheters with cerebrospinal fluid shunt infection organisms and resistance.
\newblock \emph{Journal of the Pediatric Infectious Diseases Society}, 12\penalty0 (9):\penalty0 504--512, 2023.

\bibitem[Taha et~al.(2025)Taha, Luo, Naik, Sabal, Sun, McGovern, Sandoval-Garcia, and Guillaume]{tahaAutomatedVentricularSegmentation2025}
Birra~R Taha, Gaoxiang Luo, Anant Naik, Luke Sabal, Ju~Sun, Robert~A McGovern, Carolina Sandoval-Garcia, and Daniel~J Guillaume.
\newblock Automated ventricular segmentation in pediatric hydrocephalus: how close are we?
\newblock \emph{Journal of Neurosurgery: Pediatrics}, 1\penalty0 (aop):\penalty0 1--8, 2025.

\bibitem[Yang et~al.(2025)Yang, Li, Yang, Zhang, Hui, Zheng, Yu, Gao, Huang, Lv, et~al.]{yang2025qwen3}
An~Yang, Anfeng Li, Baosong Yang, Beichen Zhang, Binyuan Hui, Bo~Zheng, Bowen Yu, Chang Gao, Chengen Huang, Chenxu Lv, et~al.
\newblock Qwen3 technical report.
\newblock \emph{arXiv preprint arXiv:2505.09388}, 2025.

\bibitem[{ZhipuAI}()]{zhipu-embedding-v3}
{ZhipuAI}.
\newblock {zhipu-embedding-v3}.
\newblock \url{https://open.bigmodel.cn/dev/api#text_embedding}.
\newblock Accessed: 2025-06-29.

\end{thebibliography}





\end{document}